\documentclass[11pt]{article}

\usepackage[a4paper,margin=1in]{geometry}
\usepackage{graphicx}
\usepackage{multirow}
\usepackage{amsmath,amssymb,amsfonts}
\usepackage{amsthm}
\usepackage{mathrsfs}
\usepackage[title]{appendix}
\usepackage{xcolor}
\usepackage{textcomp}
\usepackage{manyfoot}
\usepackage{booktabs}
\usepackage{algorithm}
\usepackage{algorithmicx}
\usepackage{algpseudocode}
\usepackage{listings}
\usepackage[numbers]{natbib}
\usepackage[hidelinks]{hyperref}

\theoremstyle{definition}

\title{Spectrally unstable nodes drive reliability failures in graph learning}

\author{
Yongyu Wang\\
ORCID: 0009-0006-0705-752X\\
Email: \texttt{yongyuw@mtu.edu}
}

\date{}

\begin{document}

\maketitle

\begin{abstract}
Graph-learning algorithms can fail when graph structure is adversarially perturbed, intrinsically noisy or constructed from imperfect observations. Here we show that some nodes bear much greater responsibility than others for allowing adversarial perturbations and intrinsic noise to harm graph-learning algorithms. Building on graph-spectral distortion analysis, we identify these failure-driving nodes and introduce a reliability-aware intervention that isolates them from the main learning step. The target algorithm is applied to a stable induced subgraph, and predictions for isolated nodes are recovered through topology- or centroid-based propagation. Across graph neural networks under targeted and non-targeted structural attacks, spectral hypergraph clustering and multi-view spectral clustering, this principle improves reliability under both adversarial and intrinsic noise. These results suggest that node-level spectral instability provides a common mechanism for understanding and mitigating reliability failures in graph learning.
\end{abstract}


\section{Introduction}\label{sec:intro}

Graph-learning methods have become a central tool for extracting information from relational data. Many real-world systems are naturally represented as graphs, where nodes denote entities and edges encode interactions, dependencies, or similarities. Examples include citation networks, social networks, transportation systems, biological interaction networks, hypergraphs with higher-order relations, and multi-view datasets with heterogeneous feature representations \cite{hu2020open,bretto2013hypergraph,chaudhuri2009multi,xu2016weighted}. By exploiting such relational structure, graph neural networks (GNNs) can propagate and aggregate information over neighbourhoods for node classification, while spectral methods use graph Laplacians to reveal cluster structure through low-dimensional embeddings \cite{wu2020comprehensive,von2007tutorial}. However, the same dependence on graph structure also creates a distinctive vulnerability. When edges are adversarially perturbed, hyperedges contain irrelevant or missing nodes, views carry inconsistent information, or graphs are constructed from noisy observations, the resulting structural errors can be amplified by message passing, spectral embedding, or graph fusion \cite{zugner2018adversarial,xu2019topology,nadler2006fundamental,zhang2022deep,pu2023robust}. 

Existing attempts to improve graph-learning reliability have mostly addressed this problem by modifying graph edges, purifying adjacency structure, or changing the learning procedure \cite{wu2019adversarial,entezari2020all,chen2025adedgedrop}. 

Here we study graph-learning reliability from a node-level vulnerability perspective. We ask, for each node in the graph, whether it is responsible for translating external adversarial perturbations and intrinsic noise into harm to the graph-learning algorithm, and how much responsibility it bears.

Our central hypothesis is that graph-learning reliability failures are disproportionately associated with spectrally unstable nodes: nodes whose local relationships undergo large distortion when the input relational structure is transformed into learned or embedded representations. This perspective builds on graph-spectral robustness analysis, in which dominant generalized spectral directions characterize large distortion between an input manifold and an output manifold \cite{cheng2021spade}. 

To operationalize this idea, we construct an input-side manifold graph from the original data relationships and an output-side manifold graph from the representations produced by the target graph-learning algorithm. The two graph Laplacians are compared through a generalized spectral analysis based on the Courant--Fischer variational characterization \cite{Golub1983MatrixC,chung1996spectral}. For each local relationship in the input-side manifold, the corresponding edge-level distortion score measures how strongly that relationship aligns with the dominant input-output distortion directions. We then convert these edge-level scores into a node-level spectral instability score by averaging the distortion scores of the local relationships incident to each node. A high node-level score indicates that the node participates strongly in the dominant distortion between the input-side relational structure and the output-side representation structure, whereas a low score indicates that the node lies in a more stable region whose local relational structure is more consistently preserved.

This node-level view leads to a reliability-aware intervention for graph learning. Rather than exposing the target algorithm to all nodes, we first identify high-instability nodes and isolate them from the main learning step. The remaining nodes are treated as the stable node set, and the target algorithm is applied to the induced subgraph formed by this set. After the target algorithm has produced reliable task outputs on the stable substructure, such as class predictions in supervised learning or cluster assignments in unsupervised learning, the isolated nodes are assigned their outputs through controlled recovery, using topology-based propagation from neighbouring stable nodes or centroid-based assignment.

The proposed framework is designed as a general reliability framework for graph learning, allowing problems in many algorithms that are usually studied separately to be examined from a unified node-level instability perspective. In GNNs, adversarial edge perturbations can redirect message passing and corrupt node representations \cite{zugner2018adversarial,xu2019topology,sun2022adversarial}. In spectral hypergraph clustering, noisy or inaccurate hyperedges can distort higher-order relational structure and degrade clustering quality \cite{bretto2013hypergraph,zhang2022deep,bulo2009game,liu2016elastic}. In multi-view spectral clustering, unreliable view-specific graphs and noisy feature representations can accumulate during graph fusion and distort the final spectral embedding \cite{chaudhuri2009multi,xu2016weighted,pu2023robust}. Although these settings differ in data structure and learning procedure, they share a common mechanism: the target algorithm depends on a graph-derived relational structure to transform input data into output representations, and adversarial perturbations or intrinsic noise harm the algorithm by disrupting the graph structure on which this transformation relies.

In this work, we evaluate this principle across three graph-learning settings. First, we study GNN reliability under targeted Nettack and non-targeted Metattack structural attacks, using multiple backbone architectures and comparing against representative graph-purification and edge-dropping defences \cite{zugner2018adversarial,zügner2024adversarialattacksgraphneural,wu2019adversarial,entezari2020all,chen2025adedgedrop}. Second, we apply the same instability-based intervention to spectral hypergraph clustering, where the main challenge is inaccurate higher-order relations. Third, we extend the framework to multi-view spectral clustering, where noise can arise from imperfect features, unreliable view-specific graphs, and accumulated errors during affinity fusion. Across these settings, isolating spectrally unstable nodes from the main learning step improves reliability relative to undefended baselines and existing graph-processing strategies. These results suggest that node-level spectral instability provides a common lens for understanding why graph-learning systems fail and how their failures can be mitigated.

Our work makes two main contributions. First, we reveal a node-level mechanism underlying the failure of graph-learning algorithms under external adversarial perturbations and intrinsic noise. Specifically, we show that such failures are not caused by all nodes equally, but are driven disproportionately by a small subset of spectrally unstable nodes that bear greater responsibility for translating structural disruptions into algorithmic errors. This shifts the focus from edge purification and global graph denoising to the concentration of failure-driving structure around particular nodes. Second, we propose a reliability-aware graph-learning intervention that isolates unstable nodes from the core learning step of the algorithm and recovers their outputs through controlled propagation. Together, these contributions provide both a node-level mechanistic interpretation of graph-learning failures caused by harmful perturbations to graph topology and a practical reliability framework for mitigating such failures.

\section{Results}

We evaluated whether spectrally unstable nodes can explain and mitigate reliability failures across graph-learning algorithms. The experiments were designed to answer three questions. First, when graph-learning failures occur under harmful perturbations to graph topology, do some nodes play a disproportionately important role in translating these perturbations into algorithmic failures? Second, can isolating these high-responsibility nodes from the core learning step improve reliability under adversarial structural perturbations and intrinsic topological noise? Third, if isolation improves reliability, does it matter which nodes are isolated?

\subsection{Reliability failures concentrate around spectrally unstable nodes}

To test whether spectral instability identifies failure-prone nodes, we first examined how attack-induced errors are distributed across instability scores. We distinguished general prediction errors from attack-induced reliability failures. A node was counted as an attack-induced failure only if it was correctly classified by a model trained and evaluated on the clean graph, but misclassified by the same model trained and evaluated on the attacked graph. This definition reduces the confounding effect of nodes that are intrinsically difficult to classify and focuses on errors induced by structural perturbation.

\begin{figure*}[!htbp]
\centering
\includegraphics[scale=0.65]{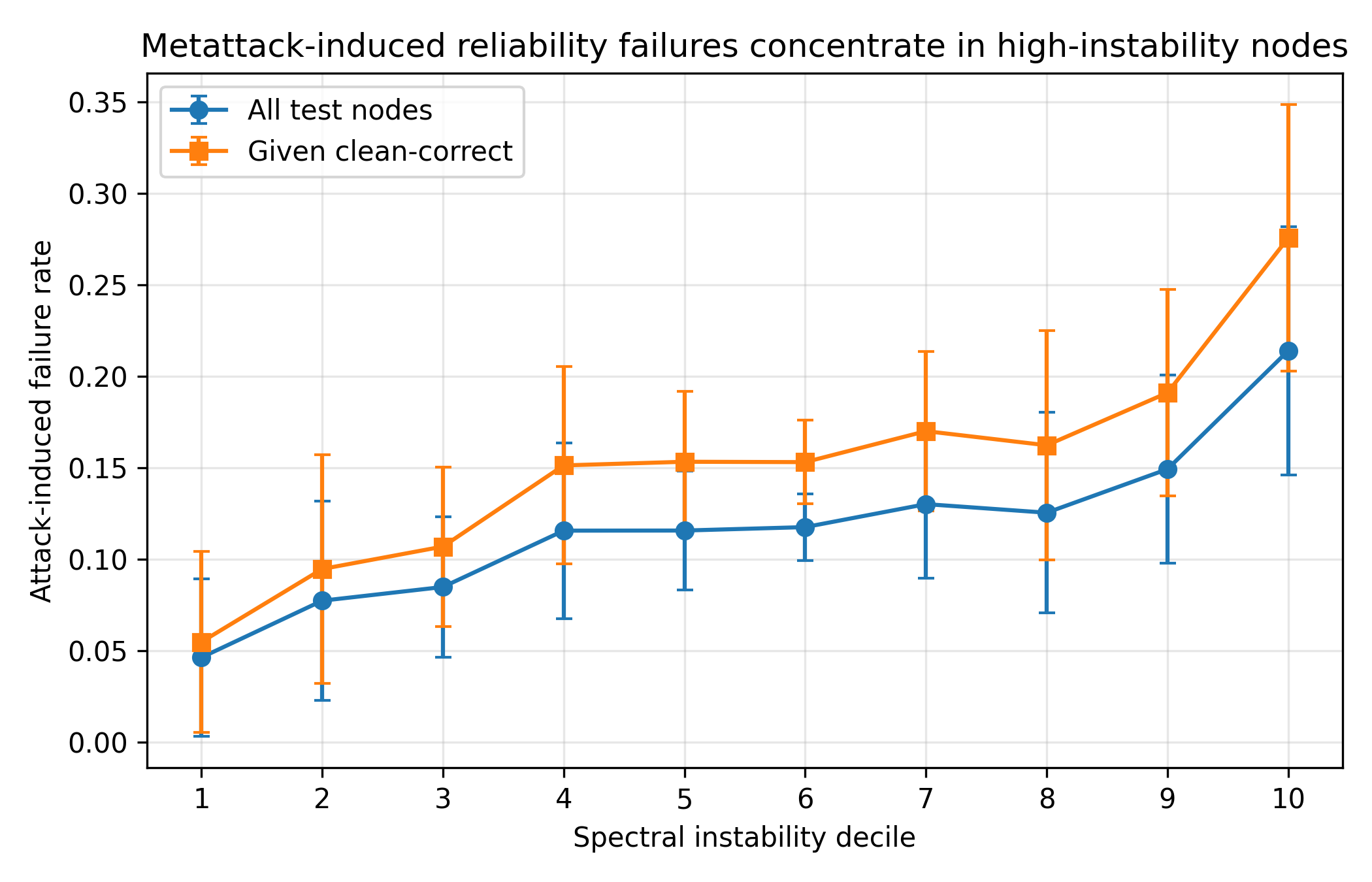}
\caption{Metattack-induced reliability failures concentrate around spectrally unstable nodes. Test nodes were sorted by spectral instability and divided into ten equal-sized deciles. Results are averaged over six adversarial GNN settings: Cora and Citeseer with GCN, GAT and GPR-GNN backbones. A node was counted as an attack-induced failure if it was correctly classified on the clean graph but misclassified under the Metattack-perturbed graph. Error bars indicate the standard deviation across the six settings. The mean attack-induced failure rate increased from 4.63\% in the lowest-instability decile to 21.40\% in the highest-instability decile, and from 5.48\% to 27.57\% when conditioned on clean-correct nodes.}
\label{fig:metattack_failure_concentration}
\end{figure*}

Using adversarial node classification with GNNs as a controlled test case, we evaluated six Metattack settings covering two citation networks and three backbone architectures. Across these settings, attack-induced failures were concentrated among high-instability nodes, as shown in Figure~\ref{fig:metattack_failure_concentration}.

Test nodes were divided into ten equal-sized groups according to spectral instability. Averaged over Cora and Citeseer with GCN, GAT and GPR-GNN backbones, the attack-induced failure rate increased from 4.63\% in the lowest-instability decile to 21.40\% in the highest-instability decile. Among nodes that were correctly classified on the clean graph, the conditional attack-induced failure rate increased from 5.48\% to 27.57\%. Thus, nodes in the highest-instability decile were approximately 4.6 times more likely to become failures under Metattack than nodes in the lowest-instability decile, and approximately 5.0 times more likely when conditioned on clean correctness.

\begin{figure}[!htbp]
\centering
\includegraphics[width=\columnwidth]{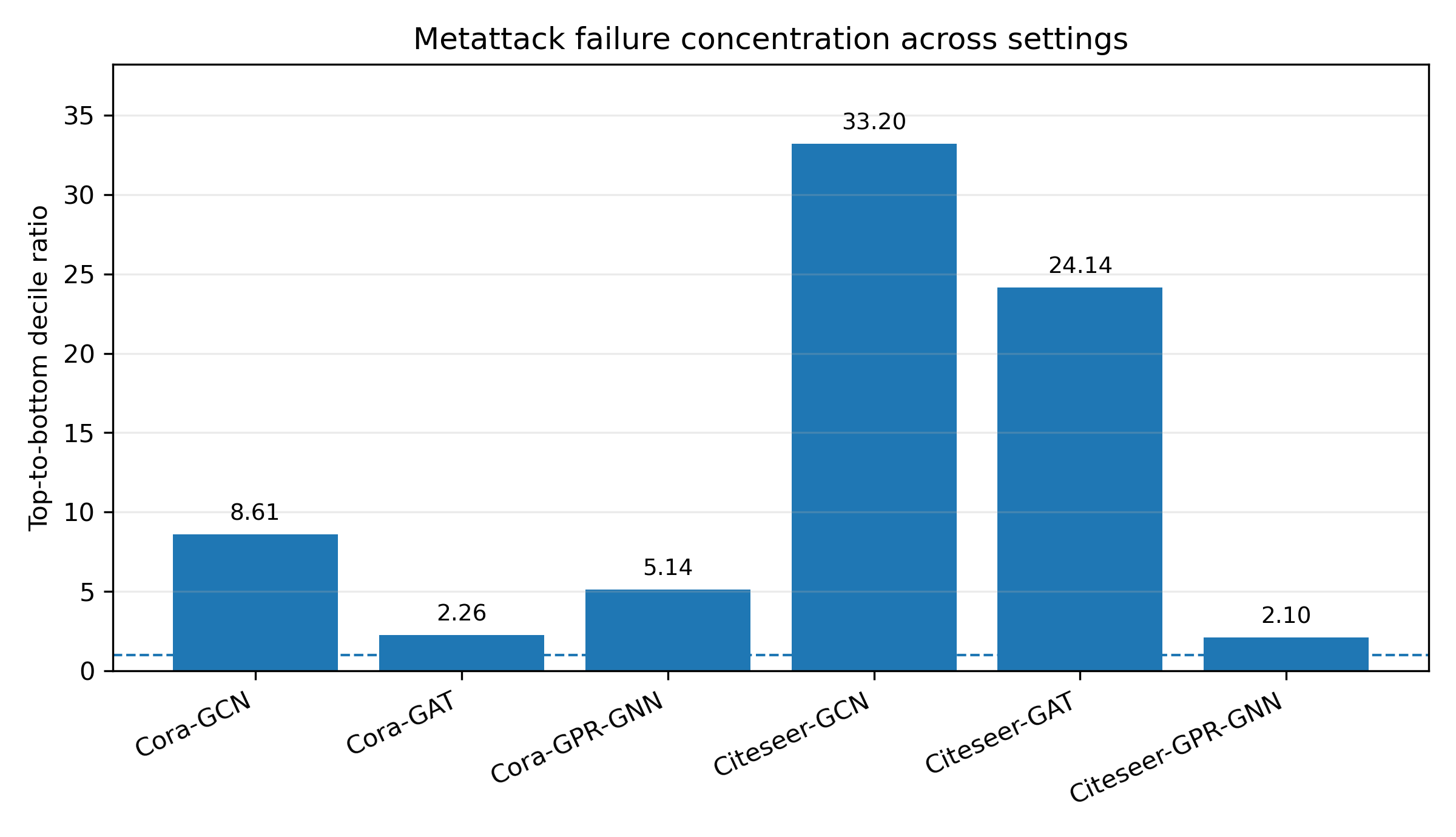}
\caption{Top-to-bottom decile ratio of Metattack-induced failure concentration across dataset--backbone combinations. The ratio is defined as the attack-induced failure rate in the highest-instability decile divided by that in the lowest-instability decile. The dashed horizontal line at 1 indicates equal failure rates in the highest- and lowest-instability deciles. All values exceed 1, showing that attack-induced failures are consistently more concentrated among high-instability nodes.}
\label{fig:metattack_decile_ratio}
\end{figure}

The same trend appeared in every evaluated Metattack setting. To quantify the strength of this concentration, we computed the top-to-bottom decile ratio, defined as the attack-induced failure rate in the highest-instability decile divided by the attack-induced failure rate in the lowest-instability decile. A ratio larger than one indicates that failures are more concentrated among high-instability nodes. As shown in Fig.~\ref{fig:metattack_decile_ratio}, this ratio exceeded one in every dataset--backbone combination: 8.61 for Cora--GCN, 2.26 for Cora--GAT, 5.14 for Cora--GPR-GNN, 33.20 for Citeseer--GCN, 24.14 for Citeseer--GAT and 2.10 for Citeseer--GPR-GNN. Thus, in every evaluated Metattack setting, nodes with the highest spectral instability were more likely to become attack-induced failures than nodes with the lowest spectral instability. These results indicate that spectral instability is not merely a descriptive graph quantity; instead, it identifies a subset of nodes that bears a disproportionately large share of perturbation-induced reliability failures across datasets and GNN architectures.

\subsection{Isolating high-instability nodes improves Cora robustness}

Having observed that attack-induced failures are enriched among spectrally unstable nodes, we next tested whether these nodes can be used as an actionable intervention target. We isolated the top 5\% highest-instability nodes from the core learning step and compared this intervention with isolating the same proportion of random nodes or low-instability nodes. The 5\% intervention was tested on Cora under Metattack using three GNN backbones: GCN, GAT and GPR-GNN. The target GNN was trained on the remaining stable induced subgraph, and predictions for isolated nodes were recovered through topology-based propagation before accuracy was evaluated on the original full test set.

\begin{figure*}[!htbp]
\centering
\includegraphics[width=\linewidth]{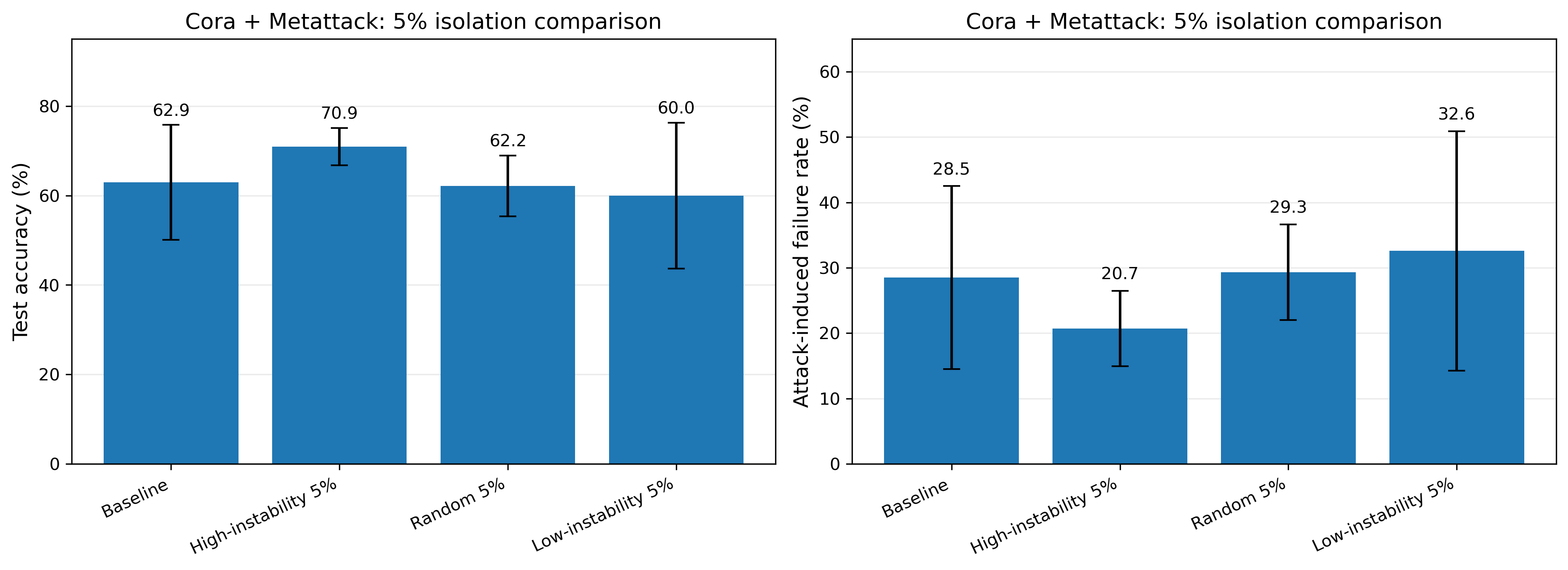}
\caption{Isolation of the top 5\% high-instability nodes improves Cora robustness under Metattack. Left, test accuracy for the attacked-graph baseline, 5\% high-instability isolation, 5\% random isolation and 5\% low-instability isolation. Right, attack-induced failure rate conditioned on nodes that were correctly classified on the clean graph. In each isolation setting, the selected nodes are removed only from the core learning step; their predictions are recovered by propagating outputs from the remaining stable nodes before evaluation. Accuracy and failure rates are therefore computed on the original full test set, including both stable nodes and recovered isolated nodes. Results are averaged over GCN, GAT and GPR-GNN backbones. Random isolation is first averaged over random trials within each backbone and then averaged across backbones. Error bars indicate the standard deviation across the three backbones.}
\label{fig:cora_fixed5_intervention}
\end{figure*}

Isolating the top 5\% high-instability nodes improved both accuracy and reliability (Fig.~\ref{fig:cora_fixed5_intervention}). Averaged over the three GNN backbones, the attacked-graph baseline achieved 62.93\% test accuracy. After isolating the top 5\% high-instability nodes, the mean test accuracy increased to 70.91\%, corresponding to a gain of 7.98 percentage points. In contrast, isolating the same proportion of random nodes reached 62.15\%, and isolating the same proportion of low-instability nodes reached 59.98\%. Thus, under the same 5\% isolation ratio, high-instability isolation produced the clearest accuracy improvement.

The same intervention also reduced attack-induced reliability failures. Among nodes that were correctly classified on the clean graph, the conditional attack-induced failure rate was 28.49\% for the attacked-graph baseline. Isolating the top 5\% high-instability nodes reduced this rate to 20.72\%, a reduction of 7.77 percentage points. By comparison, isolating the same proportion of random nodes yielded a failure rate of 29.31\%, and isolating the same proportion of low-instability nodes yielded 32.59\%. 

These results indicate that high-instability nodes provide a useful intervention target. By excluding these nodes from the core learning step and then recovering their predictions from the remaining stable nodes, the intervention substantially mitigates graph-learning failures under structural perturbation.

These findings also provide a controlled test of whether the spectral-instability ranking identifies a small set of nodes whose isolation improves reliability. The comparison with random and low-instability isolation uses the same 5\% isolation ratio, so the observed improvement cannot be attributed simply to removing nodes from the graph. Instead, the result shows that which nodes are isolated matters. Under the same 5\% isolation ratio, random isolation slightly reduced test accuracy and slightly increased the conditional attack-induced failure rate relative to the attacked-graph baseline, whereas low-instability isolation caused a larger accuracy drop and a larger increase in failure rate. In contrast, high-instability isolation substantially improved accuracy and reduced attack-induced failures, indicating that the benefit comes from targeting unstable nodes rather than from node isolation itself.

\subsection{Spectral-instability isolation improves adversarial robustness across GNN backbones}

We next evaluated whether the proposed spectral-instability intervention improves adversarial robustness across graph-learning models, datasets and attack types. We considered two representative datasets, Cora and Citeseer, three GNN backbones, GCN, GAT and GPR-GNN, and two structural attacks, the non-targeted Metattack and the targeted Nettack. For each setting, we compared the attacked-graph backbone model with three representative defense methods: adaptive edge dropping~\cite{chen2025adedgedrop}, Jaccard-based edge filtering~\cite{wu2019adversarial} and SVD-based graph purification~\cite{entezari2020all}. The proposed intervention isolates high-instability nodes, trains the target GNN on the remaining stable induced subgraph and recovers full-graph predictions through topology-based propagation.

Across all twelve dataset--model--attack combinations, the proposed intervention consistently improved over the undefended attacked-graph baseline (Table~\ref{tab:gnn_results_combined}). The average classification accuracy increased from 56.83\% for the undefended baseline to 67.43\%, corresponding to an average gain of 10.60 percentage points. The improvement was observed across all three backbone families. For GCN, the average accuracy increased from 58.35\% to 67.74\%. For GAT, the average accuracy increased from 50.85\% to 64.47\%, indicating that the intervention is especially beneficial when attention-based aggregation is strongly disrupted by adversarial edges. For GPR-GNN, the average accuracy increased from 61.29\% to 70.10\%, showing that the intervention also benefits propagation-based GNN architectures.

Compared with existing defenses, the proposed intervention achieved the highest accuracy in 9 out of 12 settings. Edge-level defenses were competitive in several cases, but their behavior was less stable across attacks and backbones. For example, ADEdgeDrop performed well under several Metattack settings, but degraded performance under targeted Nettack in some Cora settings, reducing GAT accuracy from 69.97\% to 59.61\% and GPR-GNN accuracy from 60.31\% to 43.21\%. Jaccard filtering performed strongly in several Cora settings, but was less reliable on Citeseer under Nettack, where GCN accuracy decreased from 53.61\% to 47.99\%. SVD-based purification was generally weaker, especially under Nettack, where it substantially degraded GPR-GNN on Cora from 60.31\% to 29.18\%.

These results suggest that the proposed node-level intervention provides a more stable robustness mechanism than edge-level filtering or low-rank graph approximation, and supports the view that spectrally unstable nodes are actionable drivers of reliability failures in graph learning.

\begin{table*}[!htbp]
\centering
\caption{Classification accuracy (\%) of GNN backbones under targeted Nettack and non-targeted Metattack. SI denotes spectral instability. The proposed SI intervention consistently improves over the undefended attacked-graph baseline across all twelve dataset--backbone--attack combinations and achieves the highest accuracy in 9 out of 12 combinations. The best result in each row is shown in bold and underlined.}
\label{tab:gnn_results_combined}
\footnotesize
\setlength{\tabcolsep}{5pt}
\renewcommand{\arraystretch}{1.12}
\begin{tabular}{lllccccc}
\hline
Model & Dataset & Attack & Baseline & ADEdgeDrop & Jaccard & SVD & Proposed SI intervention \\
\hline
GCN & Cora & Metattack & 56.34 & 71.13 & \underline{\textbf{71.53}} & 63.98 & 66.45 \\
GCN & Cora & Nettack & 62.42 & 72.79 & 74.60 & 65.04 & \underline{\textbf{74.75}} \\
GCN & Citeseer & Metattack & 61.02 & \underline{\textbf{68.72}} & 67.77 & 62.44 & 66.17 \\
GCN & Citeseer & Nettack & 53.61 & 61.67 & 47.99 & 37.09 & \underline{\textbf{63.57}} \\
\hline
GAT & Cora & Metattack & 42.25 & 68.41 & \underline{\textbf{70.98}} & 58.40 & 66.50 \\
GAT & Cora & Nettack & 69.97 & 59.61 & 54.48 & 60.06 & \underline{\textbf{81.89}} \\
GAT & Citeseer & Metattack & 49.59 & 52.73 & 51.84 & 52.19 & \underline{\textbf{57.35}} \\
GAT & Citeseer & Nettack & 41.59 & 46.86 & 48.99 & 37.20 & \underline{\textbf{52.13}} \\
\hline
GPR-GNN & Cora & Metattack & 69.37 & 74.75 & 73.24 & 66.15 & \underline{\textbf{76.61}} \\
GPR-GNN & Cora & Nettack & 60.31 & 43.21 & 66.75 & 29.18 & \underline{\textbf{67.81}} \\
GPR-GNN & Citeseer & Metattack & 56.93 & 67.48 & 67.48 & 60.13 & \underline{\textbf{69.02}} \\
GPR-GNN & Citeseer & Nettack & 58.53 & 65.28 & 63.86 & 41.94 & \underline{\textbf{66.94}} \\
\hline
Average & -- & -- & 56.83 & 62.72 & 63.29 & 52.82 & \underline{\textbf{67.43}} \\
\hline
\end{tabular}
\end{table*}

\subsection{Spectral-instability isolation generalizes to complex graph-clustering settings}

We next tested whether the proposed intervention is effective not only against explicit external attacks, but also against implicit and intrinsic sources of unreliability, such as intrinsic noise, imprecise data relationships and imperfectly constructed graph topology. To this end, we evaluated the proposed intervention on two complex spectral clustering settings that are naturally exposed to such problems: spectral hypergraph clustering and multi-view spectral clustering.

Standard spectral hypergraph clustering first represents higher-order relations as a hypergraph, converts the hypergraph into a clique-expansion graph and then performs spectral clustering in the resulting embedding space \cite{purkait2016clustering}. Standard multi-view spectral clustering first constructs an affinity graph for each feature view, fuses the view-specific graphs and then performs spectral clustering on the fused graph structure \cite{huang2012affinity}. These two settings expose the algorithm to different implicit sources of unreliability: inaccurate higher-order relations in hypergraphs, and unreliable feature views, graph-construction errors and accumulated fusion noise in multi-view graphs \cite{nadler2006fundamental,bulo2009game,liu2016elastic,zhang2022deep,pu2023robust}. We evaluated spectral hypergraph clustering on three hypergraph benchmarks: Zoo, Tic-Tac-Toe Endgame and Car Evaluation. We evaluated multi-view spectral clustering on six multi-view benchmarks: BBCSport, Wikipedia-test, Prokaryotic, OutdoorScene, Caltech101-7 and Handwritten.

In both settings, the proposed intervention follows the same principle. It first evaluates node-level spectral instability, identifies unstable nodes, performs the main spectral clustering step on the remaining stable substructure and then assigns the isolated nodes through controlled recovery. In spectral hypergraph clustering, recovery is performed by propagating labels from stable neighbouring nodes in the clique-expansion graph. In multi-view spectral clustering, recovery is performed by centroid-based assignment, which assigns each isolated point to the cluster whose centroid is closest in the embedding space.

Across all nine complex clustering benchmarks, spectral-instability isolation improved accuracy over the corresponding standard spectral clustering baseline (Table~\ref{tab:complex_clustering_results}). Notably, our method achieved accuracy improvements of more than 10 percentage points on five of them.

\begin{table*}[!htbp]
\centering
\caption{Clustering accuracy (\%) on complex graph-clustering benchmarks. SI denotes spectral instability. The proposed SI intervention improves over the corresponding standard spectral clustering baseline on all spectral hypergraph clustering and multi-view spectral clustering datasets.}
\label{tab:complex_clustering_results}
\footnotesize
\setlength{\tabcolsep}{6pt}
\renewcommand{\arraystretch}{1.12}
\begin{tabular}{llccc}
\hline
Setting & Dataset & Standard baseline & Proposed SI intervention & Gain \\
\hline
Hypergraph & Zoo & 62.38 & 75.25 & 12.87 \\
Hypergraph & Tic-Tac-Toe Endgame & 51.67 & 54.59 & 2.92 \\
Hypergraph & Car Evaluation & 37.62 & 50.87 & 13.25 \\
\hline
Multi-view & BBCSport & 42.34 & 55.27 & 12.93 \\
Multi-view & Wikipedia-test & 53.15 & 57.09 & 3.94 \\
Multi-view & Prokaryotic & 51.91 & 66.06 & 14.15 \\
Multi-view & OutdoorScene & 64.96 & 69.68 & 4.72 \\
Multi-view & Caltech101-7 & 54.82 & 66.82 & 12.00 \\
Multi-view & Handwritten & 79.95 & 85.05 & 5.10 \\
\hline
\end{tabular}
\end{table*}

Together, these clustering results support the broader interpretation of spectrally unstable nodes as failure-driving elements in graph learning: they distort the transformation from input relational structure to output representation, and isolating them can improve reliability across both supervised and unsupervised graph-learning tasks.

\section{Discussion}

This work identifies spectrally unstable nodes as a node-level mechanism underlying reliability failures in graph learning. Rather than treating graph-learning failures as uniformly distributed consequences of global graph noise, we show that perturbation-induced failures are concentrated around a subset of nodes whose local relational structure undergoes large spectral distortion as the algorithm transforms the input data into output representations. 

The proposed view differs from conventional graph purification strategies. Existing defences often attempt to remove or reweight unreliable edges, suppress high-rank components of the adjacency matrix or learn adaptive edge-dropping policies. These approaches operate primarily at the edge or graph level. In contrast, our framework changes the unit of reliability analysis from edges to nodes. A node is treated as unreliable when it participates strongly in the dominant spectral distortion between the input relational structure and the learned or embedded output representation. This node-level perspective enables a different intervention: the main learning process is performed on a stable induced subgraph, and predictions for isolated nodes are recovered only after the stable structure has been inferred.

The experimental results support this interpretation across several levels. First, failure-concentration analyses show that attack-induced GNN errors are enriched among high-instability nodes. Second, intervention experiments under Metattack show that isolating a small fraction of high-instability nodes substantially improves test accuracy and reduces attack-induced failures, with evaluation performed on the full test set after recovering the isolated nodes. By contrast, isolating the same number of randomly selected nodes slightly reduces accuracy and slightly increases the failure rate, whereas isolating the same number of stable nodes causes a much larger accuracy drop and a much larger increase in failure rate. Third, across twelve adversarial GNN settings, the proposed intervention consistently improves over the undefended attacked-graph baseline and achieves the highest accuracy in most settings. Finally, the same principle improves spectral hypergraph clustering and multi-view spectral clustering, suggesting that spectral instability captures a broader form of graph-learning unreliability rather than a phenomenon limited to one model family or attack type.

The generality of the framework is important because graph-learning reliability failures arise in multiple forms. In GNNs, adversarial perturbations can corrupt message passing and learned node representations. In hypergraphs, noisy or incomplete hyperedges can distort higher-order relational structure. In multi-view clustering, redundant or noisy views can introduce unreliable affinity graphs, and errors can accumulate during fusion. These settings differ in model architecture, data type and learning objective, but they share a common transformation: an input relational structure is mapped into an output representation used for prediction or clustering. Spectral instability measures where this transformation is most distorted at the node level, thereby providing a common reliability signal across tasks.

Overall, the results support a reliability-centred interpretation of graph learning: graph-learning systems fail not only because the graph contains noisy edges or perturbed structures, but because certain nodes amplify these unreliable structures during the transformation from input relations to output representations. Identifying and isolating such spectrally unstable nodes provides both an explanatory mechanism and a practical intervention for improving graph-learning reliability.

\section{Methods}

\subsection{Overview of the reliability-aware graph-learning framework}

The proposed framework consists of three main steps. First, for a target graph-learning algorithm, we compare the input-side relational structure with the output-side representation learned or produced by the algorithm. This comparison yields a node-level spectral instability score. Second, nodes with high instability are isolated from the main inference step, and the target algorithm is applied to the stable induced subgraph formed by the remaining nodes. Third, predictions or cluster assignments for the isolated nodes are recovered using topology-based propagation or centroid-based assignment in clustering tasks.

\subsection{Node-level spectral instability}

To quantify how a graph-learning algorithm deforms local relational structure, we compare an input-side relational graph with an output-side manifold graph. Let \(G_X=(V,E_X,W_X)\) denote the input-side graph and let \(G_Y=(V,E_Y,W_Y)\) denote the output-side graph constructed from the learned or embedded representations. The two graphs share the same node set \(V\), but encode relational structure before and after the graph-learning transformation. We compute their graph Laplacians \(L_X\) and \(L_Y\).

We consider a graph signal \(x\in\mathbb{R}^{|V|}\) defined over the nodes. Its variation on the input-side and output-side graphs is measured by the quadratic forms \(x^\top L_X x\) and \(x^\top L_Y x\). Because graph Laplacians can be singular, we use a stabilized output-side Laplacian
\[
B=L_Y+\epsilon I,
\]
where \(\epsilon>0\) is a small numerical regularization constant. We then define the relative graph variation as
\[
R(x)=
\frac{x^\top L_X x}
{x^\top Bx}.
\]
This quotient identifies directions whose smoothness differs strongly between the input-side relational structure and the output-side representation structure. According to the generalized Courant--Fischer variational characterization~\cite{Golub1983MatrixC}, the dominant distortion directions are obtained from the generalized eigenvalue problem
\[
L_X v=\zeta Bv.
\]
Equivalently, because \(B\) is stabilized, these directions can be computed from the leading eigenvectors of \(B^{-1}L_X\).

Let \(\zeta_1\geq\zeta_2\geq\cdots\geq\zeta_s\) denote the top \(s\) generalized eigenvalues, and let \(v_1,\dots,v_s\) be the corresponding eigenvectors. Following the idea of dominant generalized-eigenvector embeddings used in graph-spectral distortion analysis~\cite{cheng2021spade}, we form the weighted spectral distortion embedding
\[
V_s=
\big[v_1\sqrt{\zeta_1},v_2\sqrt{\zeta_2},\dots,v_s\sqrt{\zeta_s}\big].
\]
The row \(V_s(p,:)\) represents node \(p\) in the dominant input-output spectral distortion coordinates. Nodes that are close in the input-side graph but far apart in this spectral distortion space participate strongly in the deformation induced by the graph-learning transformation.

For a local relation \((p,q)\in E_X\), we define its edge-level spectral distortion as
\[
d(p,q)=\|V_s(p,:)-V_s(q,:)\|_2^2.
\]
This score measures how strongly the local relation between \(p\) and \(q\) contributes to the dominant distortion between the input-side and output-side graphs. We then convert these local relation scores into a node-level spectral instability score by averaging the distortion scores of the input-side relations incident to each node:
\[
\mathrm{SI}(p)=
\frac{1}{|N_X(p)|}
\sum_{q\in N_X(p)}d(p,q),
\]
where \(N_X(p)\) is the neighbour set of node \(p\) in \(G_X\). A larger \(\mathrm{SI}(p)\) indicates that node \(p\) participates more strongly in the dominant spectral distortion between the input relational structure and the output representation structure. We therefore treat high-\(\mathrm{SI}\) nodes as spectrally unstable nodes.

\subsection{Stable-subgraph intervention and recovery}

Given spectral instability scores for all nodes, we rank the nodes from highest to lowest instability. A chosen fraction of high-instability nodes is isolated from the main inference step. Let \(U\subset V\) denote the isolated node set, and let \(R=V\setminus U\) denote the remaining stable node set. We then construct \(G_R\) by extracting the subtopology induced by the stable nodes in \(R\). Thus, after the high-instability nodes are isolated, \(G_R\) contains only the remaining stable nodes and the graph relations among them.

For GNN experiments, \(G_R\) is induced from the attacked input graph because this is the graph used by the target GNN. The GNN backbone is trained on \(G_R\), using the training and validation nodes that remain in the stable subgraph. After inference on \(G_R\), predictions for nodes in \(U\) are recovered by topology-based propagation. Specifically, each isolated node receives the class with the largest weighted vote among its stable neighbours. If no stable neighbour is available, the baseline prediction is retained.

For spectral hypergraph clustering, the stable subgraph is induced from the clique-expansion graph of the hypergraph. Standard spectral clustering is then applied to the stable induced subgraph, and cluster labels are propagated from stable nodes to isolated nodes using weighted majority voting over stable neighbours.

For multi-view spectral clustering, the target spectral embedding is computed from the fused affinity graph. Stable points are clustered in the embedding space. Isolated points are then assigned to the nearest cluster centroid computed from the stable points. This centroid-based recovery avoids running propagation over a potentially dense fused affinity graph.

\subsection{GNN adversarial robustness experiments}

We evaluated the GNN intervention on Cora and Citeseer, two citation-network benchmarks. Nodes correspond to documents, edges correspond to citation links and node features are sparse bag-of-words vectors. We used the largest connected component and the DeepRobust \cite{li2020deeprobust} split protocol with 10\% of nodes for training, 10\% for validation and 80\% for testing.

We considered three GNN backbones: GCN, GAT and GPR-GNN. The GCN backbone used two graph convolutional layers with an 8-dimensional hidden representation. The GAT backbone used a two-layer attention architecture with four attention heads in the first layer. The GPR-GNN backbone used a two-layer MLP followed by generalized PageRank propagation with 10 propagation steps and Personalized PageRank initialization with restart probability 0.1.

We evaluated two structural attacks. Metattack was used as a non-targeted attack with a 10\% perturbation ratio. Nettack was used as a targeted structural attack. All attacked graphs were generated or loaded using the DeepRobust attack setting \cite{li2020deeprobust}. We compared the proposed spectral-instability intervention with the undefended attacked-graph backbone, ADEdgeDrop, Jaccard-based edge filtering and SVD-based graph purification. All methods were evaluated using classification accuracy on the test split.

For the main GNN intervention, the top 5\% highest-instability nodes were isolated. The input and output manifold graphs used for instability computation were constructed with \(k=10\) nearest neighbours, and the number of generalized eigenvectors was set to \(s=10\).

\subsection{Spectral hypergraph clustering experiments}

We evaluated spectral-instability isolation on three hypergraph clustering benchmarks: Zoo, Tic-Tac-Toe Endgame and Car Evaluation. Each categorical dataset was represented as a hypergraph by connecting samples that share categorical attribute values. The hypergraph was converted into a weighted graph through clique expansion, and standard spectral hypergraph clustering was used as the baseline.

For the proposed intervention, the clique-expansion graph served as the input-side graph. A spectral embedding of this graph was used to construct the output-side manifold graph. Node-level spectral instability was computed from the resulting input-output graph pair. Stable nodes were clustered using standard spectral clustering on the stable induced graph, and isolated nodes were recovered by weighted majority voting from stable neighbours. 

\subsection{Multi-view spectral clustering experiments}

We evaluated multi-view spectral clustering on six benchmark datasets: BBCSport, Wikipedia-test, Prokaryotic, OutdoorScene, Caltech101-7 and Handwritten. For each view, a nearest-neighbour affinity graph was constructed. The view-specific affinity graphs were fused by uniform weighting to obtain a single graph for spectral embedding. Standard multi-view spectral clustering on the fused graph served as the baseline.

For the proposed intervention, spectral instability was computed by comparing the fused input affinity graph with the manifold graph constructed from the spectral embedding. Stable points were selected according to low spectral instability and clustered in the embedding space. Isolated points were assigned to the nearest centroid computed from the stable clusters. 

\subsection{Evaluation metrics}

For node classification, we report test accuracy:
\[
\mathrm{Accuracy}=\frac{1}{|\mathcal{T}|}\sum_{i\in \mathcal{T}}\mathbf{1}\{\hat{y}_i=y_i\},
\]
where \(\mathcal{T}\) is the test set, \(\hat{y}_i\) is the predicted class and \(y_i\) is the ground-truth label.

For clustering, we report clustering accuracy:
\[
\mathrm{ACC}=\frac{1}{n}\sum_{i=1}^{n}\mathbf{1}\{y_i=\pi(c_i)\},
\]
where \(y_i\) is the ground-truth label, \(c_i\) is the predicted cluster assignment and \(\pi(\cdot)\) is the optimal label permutation found by the Hungarian algorithm \cite{papadimitriou1982combinatorial}.

\section*{Data availability}

All datasets used in this study are publicly available benchmark datasets. Cora and Citeseer are standard citation-network datasets available through DeepRobust and related graph-learning repositories. Zoo, Tic-Tac-Toe Endgame and Car Evaluation are public categorical benchmark datasets. The multi-view datasets used in the clustering experiments are public benchmark datasets commonly used for multi-view clustering evaluation. Processed data splits and attacked graph files can be released with the code.

\section*{Code availability}

Code for reproducing the spectral-instability scoring, stable-subgraph intervention, GNN adversarial robustness experiments, hypergraph clustering experiments and multi-view clustering experiments will be made available upon publication.

\section*{Acknowledgements}

The author thanks the developers of the open-source graph-learning, scientific-computing and machine-learning libraries used in this work.

\section*{Author contributions}

The author conceived the study, developed the method, designed and conducted the experiments, analysed the results and wrote the manuscript.

\section*{Competing interests}

The author declares no competing interests.

\bibliographystyle{unsrtnat}
\bibliography{sn-bibliography}

\end{document}